\pdfoutput=1
\documentclass[11pt,a4paper]{article}
\usepackage[hyperref]{acl2021}
\usepackage{times}
\usepackage{latexsym}

\newcommand{\mysubsection}[1]{\vspace{0.3em}
\noindent\textbf{#1}}
\usepackage{booktabs}
\usepackage[normalem]{ulem}

\usepackage{tabularx}
\newcolumntype{C}{>{\centering \arraybackslash}X}

\usepackage{graphicx}
\usepackage{tikz}
\usepackage{pgfplots}

\usepackage{comment}
\usepackage{microtype}

\usepackage{soul}
\usepackage{cellspace}
\newcommand{\hlc}[2][yellow]{{%
    \colorlet{foo}{#1}%
    \sethlcolor{foo}\hl{#2}}%
}

\pgfplotsset{
  every axis plot/.append style={line width=0.8pt},
  compat=1.17,
}

\newcommand{\autour}[3]{\tikz[baseline=(X.base)]\node [draw=black,text=#2,fill=#1,thick,rectangle,inner sep=2pt, rounded corners=3pt] (X) {#3};}

\aclfinalcopy 
\def\aclpaperid{3568} 

\title{Exploring Self-Identified Counseling Expertise in Online Support Forums}

\author{Allison Lahnala\(^1\), Yuntian Zhao\(^1\), Charles Welch\(^1\), Jonathan K. Kummerfeld\(^1\), \\ \textbf{Lawrence An\(^{2,4}\), Kenneth Resnicow\(^{3,4}\), Rada Mihalcea\(^1\), Verónica Pérez-Rosas\(^1\)}\\
\(^1\)Computer Science \& Engineering, University of Michigan\\
\(^2\)Medical School, University of Michigan \\
\(^3\)School of Public Health, University of Michigan\\
\(^4\)Center for Health Communications Research, University of Michigan \\
\texttt{\{alcllahn,clzhao,cfwelch,jkummerf,lcan,}  \\ \texttt{kresnic,mihalcea,vrncapr\}@umich.edu} \\}

\date{}

\begin{document}
\maketitle
\begin{abstract}

A growing number of people engage in online health forums, making it important to understand the quality of the advice they receive.
In this paper, we explore the role of expertise in responses provided to help-seeking posts regarding mental health.
We study the differences between (1) interactions with peers; and (2) interactions with self-identified mental health professionals.
First, we show that a classifier can distinguish between these two groups, indicating that their language use does in fact differ.
To understand this difference, we perform several analyses addressing engagement aspects, including whether their comments engage the support-seeker further as well as linguistic aspects, such as dominant language and linguistic style matching.
Our work contributes toward the developing efforts of understanding how health experts engage with health information- and support-seekers in social networks. More broadly, it is a step toward a deeper understanding of the styles of interactions that cultivate supportive engagement in online communities.

\end{abstract}

\section{Introduction}

Online social media forums play a critical role in health-related information sharing~\cite{record2018sought}. 
Health experts have noted that they can help reduce healthcare inequalities and improve access to health care, for instance by empowering coalitions of people living with chronic illness or specific disabilities~\cite{griffiths2012social}, or by providing an anonymous forum for people seeking emotional support~\cite{de2014mental}.
On the other hand, these forums elevate concerns about spreading medically inaccurate, misleading, or unsound information \cite{dominguez2015pediatric,gage2018cancer}, which has had harmful public health impacts \cite{poland2011age,nobles2019requests}.
One study concluded that health information seekers in forums such as Reddit are likely to enact suggested behaviors regardless of perceived credibility \cite{record2018sought}.
However, the researchers also noted that this openness to information could be an opportunity for experts to encourage healthy behaviors through information sharing.

In this landscape, it is critical to understand the dynamics that cultivate safe communities that benefit the health and well-being of their participants and the broader implications for health communication \cite{chou2009social}.
Health experts are thus considering social media's role in their interactions with patients and broader public health issues, and their role in engaging with the platforms~\cite{dominguez2015pediatric,nobles2019requests,nobles2020examining}.
This motivates an important research direction: understanding how experts engage with users in online platforms.
This can inform platform design, moderation decisions, and health promotion efforts by experts.

This work focuses on understanding the engagement with professionals in the domain of mental health with two main research questions: (RQ1) Do experts have distinct influences as compared to non-experts in their interactions with support-seekers in online mental health?; and (RQ2) Do the experts' behaviors reflect established counseling principles and findings regarding behaviors associated with positive counseling outcomes?
To answer these questions, we analyze responses from self-identified mental health professionals (MHP) to support-seekers in mental health and support communities on Reddit, and compare them to responses from other users who we refer to as \textit{peers}. This is an important comparison, as many peers share similar health experiences, which prior work has found is associated with higher empathic concern~\cite{hodges2010giving}. 

First, we test whether a text classifier can distinguish between responses to support seekers from MHPs and peers.
We find that it can, with 70\% accuracy (well above random chance of 50\%).
Second, we analyze comments leading to further engagement with the support-seeking posters, as existing counseling principles emphasize the importance of eliciting client engagement in expert counseling sessions~\cite{miller2012motivational,perez2018analyzing}. Third, we analyze the users' linguistic tendencies, drawing inspiration from analyses of counseling conversations, which have offered insight into counselor behaviors associated with high quality sessions grounded in existing theories from psychology and counseling research using computational methods~\cite{althoff2016large,perez2018analyzing,zhang2019finding,miller2012motivational}.

The main contributions of this work are: (1) We construct a dataset of mental health conversations from Reddit users with self-identified counseling expertise, covering a set of mental health subreddits annotated with categories denoting the type of mental health concern; (2) We develop a classifier that can distinguish between the language of MHPs and that of peers; (3) We perform an analysis of the differences in language use between MHPs and peers; and (4) We provide insight into language that leads to further engagement with support-seekers, comparing responses to peers and MHPs.

\section{Related Work}

\begin{table*}[t]
  \parbox{.24\linewidth}{
        \centering
        \small
        \begin{tabular}{lr}
            \toprule
            Subreddits & 77 \\
            Posts & 12,140 \\
            Poster Replies & 24,357 \\
            MHPs & 283 \\
            Peers & 56,701 \\
            \midrule
            \multicolumn{2}{c}{Comments} \\
            \midrule
            MHP & 9,685 \\
            Peer & 92,698 \\
            Total & 102,383 \\
            \midrule
            \multicolumn{2}{c}{Thread Length} \\
            \midrule
            Mean & 8.4 \\
            Median & 4 \\
            Max & 64 \\
            \bottomrule
        \end{tabular}
        \caption{Dataset statistics.}
        \label{tab:data_stats}
    }
    \hfill
    \parbox{.72\linewidth}{
        \small
        \centering
        \begin{tabular}{ll}
             \toprule
             \textbf{Post:} & u/peer\_user\_X \\
             \midrule
             \multicolumn{2}{p{0.9\linewidth}}{I've recently been struggling with paranoid thoughts, for which I was hospitalized for my own safety. I do not feel suicidal anymore, however everyday is a long struggle of thinking everyone is an undercover agent out to get me or keep tabs on what I'm doing. I was hoping to hear some tips and stories if anyone else has dealt with similar thoughts and overcome them? Or are they something I will have to deal with for the rest of my life? Thanks in advance}
    \\
            \midrule
             \textbf{Comment:} & u/MHP\_user
             \begin{tikzpicture}
                \tikz[baseline={([yshift=-0.4em]current bounding box.north)}] \draw[rounded corners=5pt,xshift=1.0em] (-0.4,-0.2) rectangle (0.4,0.2) node[pos=.5] {LPC};
            \end{tikzpicture}
             \\
             \midrule
             \multicolumn{2}{p{0.9\linewidth}}{Paranoid thoughts are scared thoughts, justified or not. If you ignore the specific content of the thoughts and focus on the emotional valence (scared), is there something you can do in those moments to feel safer?} \\
            \midrule
             \textbf{Poster Reply:} & u/peer\_user\_X \\
             \midrule
             \multicolumn{2}{p{0.9\linewidth}}{That's a good way of thinking about the situations as they arise. I will try to do that} \\
             \bottomrule
        \end{tabular}
        \caption{Example of an initial post, a reply from an MHP with the flair \textit{LPC} (Licensed Professional Counselor), and a reply from the original user.}
        \label{tab:data_sample4}
    }
\end{table*}

Studies within the education and health domains have shown that advice and help-seeking interactions in online communities contribute positively to users' well-being, learning, and skills development~\cite{Campbell16,wang2015eliciting}. This is particularly true
for applications such as computer programming, career development, mentoring, coping with chronic or life-threatening diseases, and mental health issues~\cite{baltadzhieva-chrupala-2015,Tomprou19,wang2015eliciting,de2014mental}.

In the mental health domain, studies have explored online support communities and many have found positive outcomes associated with anonymity, perceived empathy, and active user engagement~\cite{de2014mental, rheingold1993virtual, hodges2010giving, welbourne2009supportive, nambisan2011information}. Computational approaches have aided studies in mental health forums, helping reveal positive relationships between linguistic accommodation and social support across subreddits~\cite{sharma2018support}.
One example of insights from this work is that topic-focused communities like subreddits may enable more peer-engagement than non-community based platforms~\cite{sharma2020engagement}. Other studies have revealed certain trade-offs of online support platforms, such as disparities in the level of support offered toward support-seekers of various demographics~\cite{wang2018s,nobles2020examining} and in condolences extended across different topics of distress~\cite{zhou2020condolences}. Studying MHP behaviors in such scenarios might help develop approaches that balance these trade-offs.
 
Computational approaches applied in these forums have also shed light on population-level health trends and health information needs, with examinations into how depression and post-traumatic stress disorder (PTSD) affect different demographic strata~\cite{amir2019mental}.
Data mining has also been applied to understand adverse drug reactions~\cite{wang2014sideeffectptm} and public reactions towards infectious diseases~\cite{park2017tracking}. \citet{nobles2018std} highlighted the potential for these forums to aid targeted health communication, for example by sharing information in r/STD, a subreddit about sexually transmitted diseases. Another case study of r/STD revealed the prevalence of diagnoses requests, and suggested that health professionals could partner with social media platforms to positively influence crowd-sourced diagnoses and help mitigate harmful misdiagnoses~\cite{nobles2019requests}. \citet{record2018sought} found that health information seeking Reddit users are likely to enact suggested behaviors regardless of perceived credibility, providing further reason for health expert engagement to intervene when harmful information sharing occurs and promote healthy behavior.
 
 Fewer studies have analyzed expert interactions in online forums. A study in a large Q\&A community found that experts are more likely to provide help than peers and that their participation in discussions resulted in increased length and substance of discussions~\cite{Procaci17}.
Recent studies have compared interactions with experts to interactions with peers in broader scientific communities~\cite{park2020trust} and r/AskDocs on Reddit~\cite{nobles2020examining}.
The latter paper closely relates to our study, as they also consider posts from experts on Reddit, but solely within r/AskDocs about different health topics and with users of varying demographics.

The insights discussed above motivate investigations into how health experts and other users promote scientifically sound advice and offer supportive responses to health information seekers in online forums. In this work, we aim to contribute additional insights into expertise influence in online mental health communities by studying the dynamics of the communication process between support seekers and support providers.

\section{Data Collection}\label{sec:dataset}

We seek to understand the tendencies of users with professional experience, and more specifically counseling expertise, when interacting with support-seekers in online mental health and support-related forums.
In uncovering which tendencies are associated with expertise, we enable further investigation into their role in the social dynamics of online support-seeking interactions, and potential applications of insight-driven recommendations for moderators and users of these forums.

\begin{figure*}
    \centering
\begin{tikzpicture}
\begin{axis}[
    width=\linewidth,
    height=3.5cm,
    grid=both,
    bar width=.25cm,
    legend style={font=\small},
    legend pos=north west,
    symbolic x coords={health, tentat, work, anx, you, differ, discrep, they, foc.future, interrog, cogproc, quant, insight, risk, relig, assent, sad, motion, number, nonflu, we, ppron, time, leisure, ingest, family, body, home, anger, friend, filler, shehe, informal, foc.past, female, sexual, male, I, netspeak, swear}, 
    xtick=data,
    enlarge x limits=0.01,
    x tick label style={rotate=60, anchor=east, font=\footnotesize},
    ytick={0.8, 1.0, 1.2, 1.4, 1.6, 1.8, 2.0, 2.2},
    y tick label style={anchor=east, font=\footnotesize}]
    \addplot[ybar,fill=lightgray,ybar legend] coordinates  {
    	(health, 0.7959860332257565)
    	(tentat, 0.8055825728949596)
    	(work, 0.8110132828134121)
    	(anx, 0.8111684471647727)
    	(you, 0.8210292413944864)
    	(differ, 0.8553742434797355)
    	(discrep, 0.8692382037337468)
    	(they, 0.8751238940358339)
    	(foc.future, 0.8915196663317453)
    	(interrog, 0.8989293471898235)
    	(cogproc, 0.9000317012729139)
    	(quant, 0.9003896144608834)
    	(insight, 0.9057980632031888)
    	(risk, 0.9078690597633751)
    	(relig, 1.1016018716659848)
    	(assent, 1.1058051745294315)
    	(sad, 1.111442609622524)
    	(motion, 1.1147313092899014)
    	(number, 1.123251240900814)
    	(nonflu, 1.1346822371433698)
    	(we, 1.1514960646135939)
    	(ppron, 1.1743292153901252)
    	(time, 1.1760614995095569)
    	(leisure, 1.1907022197071266)
    	(ingest, 1.268569982423503)
    	(family, 1.3287365916478084)
    	(body, 1.332310541479952)
    	(home, 1.3387387079172697)
    	(anger, 1.352991457094508)
    	(friend, 1.4228051728743212)
    	(filler, 1.4910379699677137)
    	(shehe, 1.506362829049222)
    	(informal, 1.5087378339468138)
    	(foc.past, 1.519990107618276)
    	(female, 1.553857654748661)
    	(sexual, 1.575562715003554)
    	(male, 1.6370479733076038)
    	(I, 1.7324946157824643)
    	(netspeak, 1.8811274256235888)
    	(swear, 2.1973744687818817)
    }; \addlegendentry{Peers}
    \addplot[ybar,fill=white,ybar legend] coordinates  {
    	(health, 0.7959860332257565)
    	(tentat, 0.8055825728949596)
    	(work, 0.8110132828134121)
    	(anx, 0.8111684471647727)
    	(you, 0.8210292413944864)
    	(differ, 0.8553742434797355)
    	(discrep, 0.8692382037337468)
    	(they, 0.8751238940358339)
    	(foc.future, 0.8915196663317453)
    	(interrog, 0.8989293471898235)
    	(cogproc, 0.9000317012729139)
    	(quant, 0.9003896144608834)
    	(insight, 0.9057980632031888)
    	(risk, 0.9078690597633751)
    }; \addlegendentry{MHPs}
    \addplot[draw=blue,ultra thick,smooth] 
    coordinates {(health, 1)(tentat, 1)(work, 1)(anx, 1)(you, 1)(differ, 1)(discrep, 1)(they, 1)(foc.future, 1)(interrog, 1)(cogproc, 1)(quant, 1)(insight, 1)(risk, 1)(relig, 1)(assent, 1)(sad, 1)(motion, 1)(number, 1)(nonflu, 1)(we, 1)(ppron, 1)(time, 1)(leisure, 1)(ingest, 1)(family, 1)(body, 1)(home, 1)(anger, 1)(friend, 1)(filler, 1)(shehe, 1)(informal, 1)(foc.past, 1)(female, 1)(sexual, 1)(male, 1)(I, 1)(netspeak, 1)(swear, 1)};
    \end{axis}
    \node[align=center,rotate=90,font=\small] at (-1.15cm, 0.9cm) {Dominance};
    \node[align=center,rotate=90,font=\small] at (-0.8cm, 0.9cm) {(peer/MHP)};
    \end{tikzpicture}
    \caption{LIWC category dominance scores, computed as the relative use by peers divided by the relative use by MHPs, so that equal use is at $y=1$ (blue line), higher dominance by peers at $y>1$ (grey bars) and higher dominance by MHPs at $y<1$ (white bars). Showing categories where frequency of use differs by at least 10\%.}
    \label{fig:liwc}

\end{figure*}
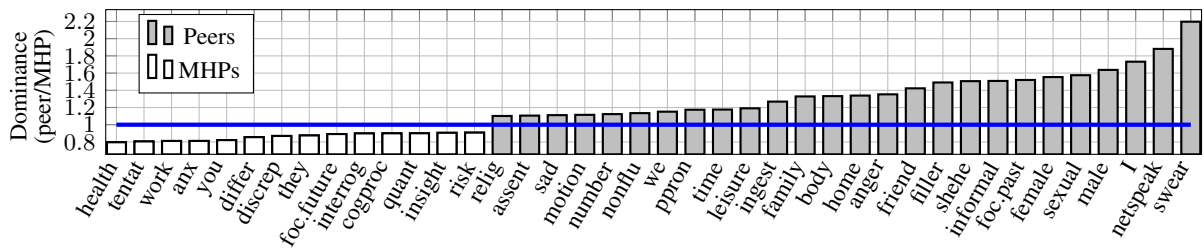

\mysubsection{Source.}
We use Reddit for its quantity of publicly available interactions in communities called \textit{subreddits} that discuss mental health issues. 
In addition, Reddit has a system that allows users to indicate their professional expertise (Reddit Flairs), which we use to identify a set of users with mental health professional background, identified as \textit{MHPs} during our study.
We obtained flairs from the r/psychotherapy subreddit,\footnote{\href{https://old.reddit.com/r/psychotherapy/wiki/acronyms\#wiki_degrees_and_licenses}{Degree and license flair descriptions from r/psychotherapy wiki.}} a decision motivated by their reliability, as the moderators of this community allow comments and posts only by licensed therapy providers who may be asked to submit proof if concerns of falsely posing as a therapist arise.\footnote{See rule 2 and 9 in \url{https://www.reddit.com/r/psychotherapy/}, also listed in Appendix~\ref{sec:appendix_flairs}.} 
Sample flair tags in this set are: Psychiatrist (sometimes accompanied by MD or DO), LPC (or Licensed Professional Counselor), LMFT (or Licensed Marriage and Family Therapist), PsyD (or Doctorate of Psychology).

We use an existing list of mental health subreddits from r/ListOfSubreddits\footnote{\href{https://www.reddit.com/r/ListOfSubreddits/comments/dmic6o/advice_mental_health_subreddits/}{r/ListOfSubreddit's compilation of mental health and advice subreddits.}}
with additions from manual observations; all of the subreddits in our dataset with their number of comments are in Appendix~\ref{sec:appendix_datacollection} in Table~\ref{tab:appendix_subbreddits_and_comments}. 
From these, we retrieve threads where an MHP submitted a direct reply. During this step, we also kept posts made by peers i.e., individuals who did not use any of the mental health care professional flairs. Our collection spans threads created between November 29, 2009 and December 21, 2020. Table~\ref{tab:data_stats} shows descriptive statistics for the final composition of the dataset, and  Table~\ref{tab:data_sample4} shows a sample interaction demonstrating the structure we use for our analysis. This study focuses on direct replies to the poster, thus we attempt to eliminate \textit{megathreads} which tend not to focus in individual support-seekers by removing those above the 95th percentile in their number of direct replies; we leave analysis of deeper nested replies for future work.

\begin{table}[t]
    \centering
    \small
    \begin{tabular}{ll}

\toprule
Key & Topic \\
\midrule
Trauma & Trauma \& Abuse \\
Anx & Psychosis \& Anxiety \\
Compuls. & Compulsive Disorders \\
Cope & Coping \& Therapy \\
Mood & Mood Disorders \\
Addict. & Addiction \& Impulse Control \\
Body & Eating \& Body  \\
Neurodiv. & Neurodevelopmental Disorders  \\
Health & General \\
Social & Broad Social \\
\bottomrule
    \end{tabular}
    \caption{Health condition and other subreddit topics. Keys are shortened names we use to refer to the topics.}
    \label{tab:subreddit_health_categories}
\end{table}

\mysubsection{Health Topics.} To understand whether particular topics influence interactions with support-seekers, we group the subreddits into broader topics based on related health domains.
We begin by following the categorization of subreddits by \citet{sharma2018support}, who used the $k$-means clustering algorithm to generate initial clusters on the $n$-grams $(n=3)$ of the posts and manually refined the categories based the community descriptions in their subreddit home pages.
Next, we adjust the categories and their associated subreddits based on the World Health Organization's ICD-10 classification system of mental and behavioural disorders.\footnote{\url{https://www.who.int/substance_abuse/terminology/icd_10/en/}} The resulting topic categories are listed in Table~\ref{tab:subreddit_health_categories} alongside shortened names which we use to refer to them. The full list of subreddits assigned to each topic are listed in Appendix~\ref{sec:appendix_datacollection} in Table~\ref{tab:appendix_subreddit_categories}.

\section{Distinguishing MHPs and Peers}\label{sec:classification}

To begin our investigation into the linguistic behaviors of MHPs and peers, we test whether simple text classifiers are able to distinguish between comments authored by either MHPs or peers.
We build three classifiers with different feature sets; the first are unigram counts for unigrams occurring at least five times, the second includes counts for the 73 word classes in the LIWC (Linguistic Inquire and Word Count) lexicon~\cite{pennebakerlinguistic}, and the third encodes a subset of LIWC word classes associated with perspective shifts (i.e., \textit{focusfuture, focuspast, focuspresent, I, ipron, negemo, posemo, ppron, pronoun, shehe, they, we,} and \textit{you})~\cite{althoff2016large}; we elaborate on the psychological meaning behind these features in our analyses in the next section.

Due to the class imbalance between the peer and MHPs classes, we first downsampled the peer class to get a balanced distribution with the MHP class. This resulted in a set of 9,685 instances per class. We conduct our evaluations using ten-fold cross validation. Across these folds, the number of features ranges from 8,668 to 8,703.
We use a Naive Bayes model, implemented with Sklearn’s MultinomialNB module,~\footnote{\url{https://scikit-learn.org/stable/modules/generated/sklearn.naive_bayes.MultinomialNB.html}} which outperformed a logistic regression model and an SVM in preliminary experiments.\footnote{Runs in $\sim$40 seconds per fold on one AMD Ryzen 7 3700U CPU.} 

All models outperform a random baseline\footnote{$p<0.0001$ using a permutation test~\cite{dror-etal-2018-hitchhikers2}} with all LIWC features bringing the accuracy to 59.12\%, LIWC perspective features to 59.14\%, and unigram features to 70.80\%.
Overall, the classification results indicate language differences exist between the MHPs and peers. Motivated by this result, we proceed to several analyses to gain insights.

\section{Linguistic and Dialogue Analysis}\label{sec:ling_and_dialog}

We analyze the linguistic behaviors of MHPs and peers responding to support-seeking posts, and their potential influence in eliciting further engagement with the support-seeker.
Our analyses are inspired by psychology and computational studies that have shown that conversational behavioral aspects such as word usage, client engagement, and language matching are positively related to successful counseling interactions~\cite{gonzales2010language,althoff2016large,perez2018analyzing,zhang2019finding}.

\begin{figure}[t]
    \centering
\begin{tikzpicture}
\begin{axis}[
    width=8.25cm,
    height=3cm,
    grid=both,
    symbolic x coords={Disgust, Surprise, Anger, Fear, Sadness, Joy},
    xtick=data,
    x tick label style={rotate=0, font=\footnotesize}, 
    ytick={0.9, 1.1, 1.3, 1.5, 1.7},
    y tick label style={anchor=east, font=\footnotesize}]
    \addplot[ybar,fill=lightgray] coordinates  {
    	(Disgust, 1.71028997750357)
    	(Surprise, 1.3983259978898)
    	(Anger, 1.14086998542706)
    	(Fear, 1.06922731579117)
    	(Sadness, 1.06749537047073)
    	(Joy, 1.00660420378226)
    }; 
\end{axis}
\node[align=center,rotate=90,font=\small] at (-0.8cm, 0.7cm) {Dominance};
\end{tikzpicture}
    \caption{WordNet Affect usage (peers / MHPs)}
    \label{fig:wordnet_affect}
\end{figure}
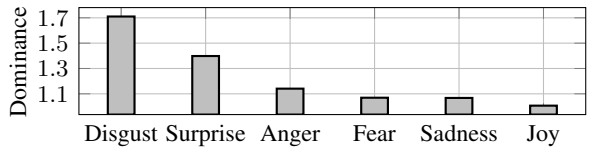

\subsection{Linguistic Ethnography}\label{sec:linguistic_ethno}

Numerous studies have demonstrated relationships between the dominant usage of certain word categories with 
individuals' psychological and physical health~\cite{tausczik2010psychological,weintraub1981verbal,rude2004language}. 
In alignment with these studies, we investigate the usage of such word categories using the LIWC and WordNet-Affect lexicons~\cite{pennebakerlinguistic,wordnet-affect}.

For each group of users, we first compute the proportion of their words that fall in each category. Then, we compute the dominant use by dividing the proportion for peer users over the proportion for MHPs~\cite{mihalcea2009linguistic}.
Figure~\ref{fig:liwc} shows 
LIWC categories where the rate of use differs by at least 10\%, and results for WordNet Affect categories are shown in Figure~\ref{fig:wordnet_affect}.

Some observations such as the higher dominance of swear words (\textit{swear}) and internet speak (\textit{netspeak}) might be expected if professionals avoid such language. An interesting contrast in peers' language is the dominant use of first-person pronouns (\textit{I, we}) and focus on the past (\textit{focuspast}). In contrast, MHPs seem to use more non-first person pronouns (\textit{you, they}) and focus on the future (\textit{focusfuture}) instead.  Peers' use of first person pronouns might arise when they share similar experiences with support-seekers. MHPs' use of second-person pronouns might suggest they are focusing on the support-seekers' experiences as a counselor would with a client in a counseling encounter.
We also observe higher dominance of all WordNet Affect categories among peers, however the \textit{joy} category (the most positive), is nearly equal with MHPs.

These observations of the peers' language are compelling because they align with existing theories linking depression to negative views of the future (i.e., \textit{focuspast} and negative WordNet affects)~\cite{pyszczynski1987depression} and self-focusing style (i.e., first-person pronouns)~\cite{pyszczynski1987self,campbell2003secret}. Likewise, clients of SMS-based crisis counseling conversations were more likely to report feeling better after the encounter if they exhibited perspective shifts from these categories to their counterparts (i.e., toward \textit{focusfuture}, non-first person pronouns, and positive sentiment)~\cite{althoff2016large}.

Interestingly, the same study found clients were more likely to shift perspective when their counselors exhibited use of the counterpart categories first, suggesting that the counselors may play a key role in helping drive the perspective shift. Given those positive outcomes, observing the same dominant linguistic aspects among MHPs is encouraging and potentially signals a connection between how counselors apply conversational behaviors in practice and in online forum interactions. Future work can investigate the progression of dialogue between MHPs and support-seekers to find if support-seekers similarly exhibit the perspective shifts associated with the positive outcomes of the prior study, and likewise whether users of the forums also experience positive outcomes where this occurs.

\begin{figure*}[t]
   \parbox{.53\linewidth}{
        \centering
        \small
        \begin{tikzpicture}
\begin{axis}[
    width=1.1\linewidth,
    height=3.7cm,
    grid=both,
    symbolic x coords={filler, health, body, bio, cause, tentat, you, affect, time, nonflu, certain, netspeak, hear, affiliation, foc.past, female, number, shehe, friend, leisure, male, ingest, anger, feel, sexual, informal, i, swear, assent, relig},
    xtick=data,
    enlarge x limits=0.01,
    bar width=.15cm,
    x tick label style={rotate=60, anchor=east, font=\small},
    ytick={0.5, 0.6, 0.7, 0.8, 0.9, 1.0, 1.1, 1.2, 1.3},
    y tick label style={anchor=east, font=\small}]
    \addplot[ybar,fill=white] coordinates  {
        (filler, 1.3694959488399698)
        (health, 1.2199518622464156)
        (body, 1.1963945578342048)
        (bio, 1.1361700435532724)
        (cause, 1.0795619328576351)
        (tentat, 1.0684715763913766)
        (you, 1.0647948805231773)
        (affect, 0.952284583106825)
        (time, 0.9517636175949772)
        (nonflu, 0.9511736159516474)
        (certain, 0.9483112140262707)
        (netspeak, 0.9463188721091061)
        (hear, 0.9389897621251143)
        (affiliation, 0.9356603838607868)
        (foc.past, 0.932086788032833)
        (female, 0.9316496264858395)
        (number, 0.9205180739190895)
        (shehe, 0.9183051123944717)
        (friend, 0.8982679230950917)
        (leisure, 0.8969320017601914)
        (male, 0.890290971657498)
        (ingest, 0.8876147841875732)
        (anger, 0.8792607168321827)
        (feel, 0.8702617564065401)
        (sexual, 0.8668272078839886)
        (informal, 0.8624389844393203)
        (i, 0.8066956501832807)
        (swear, 0.7870234652664857)
        (assent, 0.7822245598953113)
        (relig, 0.6755071110674885)
    };
    \addplot[draw=blue,ultra thick,smooth]
    coordinates {(filler, 1.0)(health, 1.0)(body, 1.0)(bio, 1.0)(cause, 1.0)(tentat, 1.0)(you, 1.0)(affect, 1.0)(time, 1.0)(nonflu, 1.0)(certain, 1.0)(netspeak, 1.0)(hear, 1.0)(affiliation, 1.0)(foc.past, 1.0)(female, 1.0)(number, 1.0)(shehe, 1.0)(friend, 1.0)(leisure, 1.0)(male, 1.0)(ingest, 1.0)(anger, 1.0)(feel, 1.0)(sexual, 1.0)(informal, 1.0)(i, 1.0)(swear, 1.0)(assent, 1.0)(relig, 1.0)};
\end{axis}
\node[align=center,rotate=90,font=\small] at (-1.0cm, 1.0cm) {Dominance Score};
\end{tikzpicture}
        
    }
    \hskip1em
    \parbox{.39\linewidth}{
        \small
        \centering
        \begin{tikzpicture}
\begin{axis}[
    width=1.2\linewidth,
    height=3.7cm,
    grid=both,
    symbolic x coords={you, foc.future, interrog, shehe, filler, male, health, nonflu, affiliation, motion, anger, home, number, informal, certain, leisure, assent, netspeak, swear, relig, i, friend, we, death},
    xtick=data,
    enlarge x limits=0.01,
    bar width=.15cm,
    x tick label style={rotate=60, anchor=east, font=\small},
    ytick={0.7, 0.8, 0.9, 1.0, 1.1},
    y tick label style={anchor=east, font=\small}]
    \addplot[ybar,fill=lightgray] coordinates  {
        (you, 1.1056504063587804)
        (foc.future, 1.0832136349699277)
        (interrog, 1.0690181983665104)
        (shehe, 1.0677545462122848)
        (filler, 1.0562119255014193)
        (male, 1.052349793324354)
        (health, 1.0510277322615278)
        (nonflu, 0.9500031679667199)
        (affiliation, 0.9457110526134682)
        (motion, 0.9432732829160265)
        (anger, 0.9416776171900192)
        (home, 0.9377753113453201)
        (number, 0.9338331405276905)
        (informal, 0.9324061632862549)
        (certain, 0.9316400785214803)
        (leisure, 0.927558556791122)
        (assent, 0.9206957666641701)
        (netspeak, 0.9202391071079117)
        (swear, 0.9141919663144525)
        (relig, 0.9118338999145017)
        (i, 0.9117467370956133)
        (friend, 0.8866945520809406)
        (we, 0.8279335154078911)
        (death, 0.8171034372572407)
    };
    \addplot[draw=blue,ultra thick,smooth]
    coordinates {(you, 1.0)(foc.future, 1.0)(interrog, 1.0)(shehe, 1.0)(filler, 1.0)(male, 1.0)(health, 1.0)(nonflu, 1.0)(affiliation, 1.0)(motion, 1.0)(anger, 1.0)(home, 1.0)(number, 1.0)(informal, 1.0)(certain, 1.0)(leisure, 1.0)(assent, 1.0)(netspeak, 1.0)(swear, 1.0)(relig, 1.0)(i, 1.0)(friend, 1.0)(we, 1.0)(death, 1.0)};
\end{axis}
\end{tikzpicture}
    }
    \caption{Dominance of LIWC categories, computed as the category relative frequencies among comments that \textbf{prompt support-seeker responses} divided by the relative frequencies among comments that do not, computed separately for MHPs (left) and peers (right).}
    \label{fig:engagement_dominance}
\end{figure*}
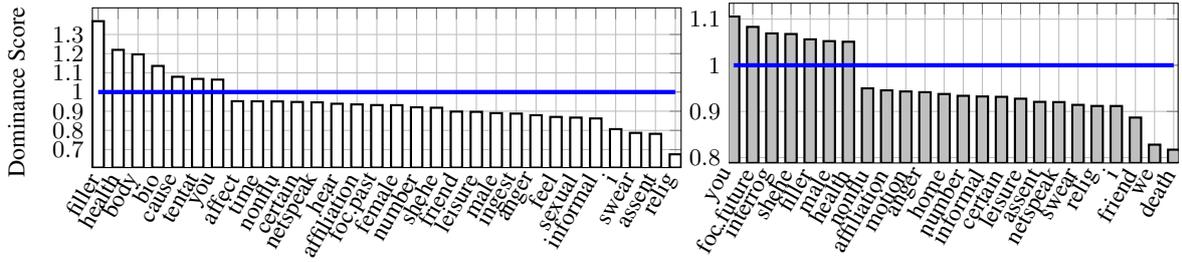
\subsection{Engaging Support-Seekers}\label{sec:engagement}

To understand if linguistic behaviors are associated with prompting further engagement with the support-seeker, we compare the dominance of LIWC categories in comments receiving replies compared to comments that do not by dividing the usage rates of the former by the latter. Figure~\ref{fig:engagement_dominance} shows these ratios for categories that differ by at least 5\%. A compelling observation is the dominance of the categories \textit{health}, \textit{tentat}, and \textit{you} in the MHP comments prompting poster-replies, and \textit{you}, \textit{focusfuture}, \textit{interrog}, and \textit{health} in the peer comments prompting poster-replies, as was exhibited among MHPs (see Figure~\ref{fig:liwc} in Section~\ref{sec:linguistic_ethno}); on the other end, the categories are more dominant in comments that do not engage a reply such as \textit{I}, \textit{we}, \textit{death}, \textit{friend}, \textit{relig}, \textit{swear}, were similarly represented as dominant categories in the peer group.

To gain further insight into these observations, we perform the following analysis: for each user group (peers and MHPs) we use a foreground corpus of their comments that were replied to by the support-seekers, and a background corpus of their comments that were not, and compute the dominance of LIWC categories of the foreground over the background as a ratio of their relative frequencies. 
We then rank the categories by highest to lowest dominance scores, and refer to this ranking by \texttt{DRR} (for \textbf{D}ominance \textbf{R}ank for \textbf{R}eplied comments). We compare the \texttt{DRR}s of each user group to the ranking of LIWC category usage among MHP users and among peer users separately (from Section~\ref{sec:linguistic_ethno}) by computing the Kendall Tau's coefficient between them.
A positive correlation would thus indicate that the more (or less) dominant categories among a group's replied comments are also more (or less) dominant among the other group overall.
The correlation coefficients are shown in Table~\ref{tab:ranked_comparison}.

\begin{table}
    \centering
    \small
    \begin{tabular}{llrr}
        \toprule
        \texttt{DRR} Group & OR Group &  $\tau$ & p-value \\
        \midrule
        Peer & MHP & .191 & .017 \\
        MHP & MHP & .158 & .048 \\
        Peer & Peer & -.031 & .689 \\
        MHP & Peer & .008 & .916 \\
        \bottomrule
    \end{tabular}
    \caption{Kendall $\tau$'s coefficient between the LIWC category dominance ranking in the replied comments (\texttt{DRR}) of the user group on the left and the overall ranking of LIWC category usage (\texttt{OR}) by the user group to the right.
    }
    \label{tab:ranked_comparison}
\end{table}

Interestingly, we observe a slight positive correlation between the \texttt{DRR}s for both MHPs and peers with the overall LIWC category usage ranking for MHPs. On the other hand, we see no correlations with the LIWC usage ranks for peers. Intuitively, it appears that for both MHPs and peers, the comments prompting further engagement with the poster appear to reflect the overall dominant linguistic aspects captured by LIWC of MHPs, but not peers. As counseling principles have emphasized the importance of mutual engagement between counselors and clients~\cite{miller2012motivational} and other work has shown that higher quality counseling sessions are associated with higher client engagement~\cite{perez2018analyzing}, it is compelling to observe associations between linguistic aspects of MHPs with the aspects associated with poster-engagement.

\subsection{Linguistic Style Matching}\label{sec:lsm}

Linguistic Style Matching (LSM) measures the extent to which one speaker matches another~\cite{gonzales2010language}. It compares two parties' relative use of function words as these words are more indicative of style rather than content~\cite{ireland2010language}.

Previous studies in counseling conversations have measured LSM to understand the extent that counselors and clients match their language. \citet{perez2019makes} showed higher LSM for high quality counseling sessions whereas \citet{althoff2016large} showed lower LSM for higher quality sessions. \citet{perez2019makes} attributed this to the differences between the conversations they analyzed, theirs being synchronous face-to-face interactions while \citet{althoff2016large}'s was of asynchronous text messages, as well as differences in counseling styles.

We follow \citet{nobles2020examining}'s approach leveraging \citet{ireland2010language}'s procedure to measure LSM between support seekers and support providers.

\begin{figure*}
    \centering
      \begin{tikzpicture}
      \begin{axis}[
      width  = \linewidth-2cm,
      height = 3cm,
      ybar=2*\pgflinewidth,
      ylabel={LSM Score},
      ylabel style={font=\small},
      ymajorgrids = true,
      symbolic x coords={Trauma,Anx,Compuls.,Cope,Mood,Addict.,Body,Neurodiv.,Health,Social,All},
      x tick label style={rotate=0, font=\footnotesize},
      xtick = data,
      scaled y ticks = false,
      ymin=0.4,
      legend cell align=left,
    legend style={at={(1.01,0.4)},anchor=west}
      ]
      \addplot[style={fill=white},error bars/.cd, y dir=both, y explicit]
          coordinates {
(Trauma,0.6310119156541756) += (0,0.029962442897110475) -= (0,0.03081514525447937)
(Anx,0.5997199013986588) += (0,0.01519174511662591) -= (0,0.015544352706185927)
(Compuls.,0.6281754095486042) += (0,0.0357519342555046) -= (0,0.03738872492676926)
(Cope,0.6473266495067116) += (0,0.010066984978184701) -= (0,0.010552935538052588)
(Mood,0.6437280427655686) += (0,0.009180262091117664) -= (0,0.009328442085683064)
(Addict.,0.5709948285212645) += (0,0.015199256087097268) -= (0,0.015134390184438407)
(Body,0.5887944659204036) += (0,0.04432956318398418) -= (0,0.04696878984370534)
(Neurodiv.,0.6309063259324916) += (0,0.01391890784736749) -= (0,0.014303173962970095)
(Health,0.6144863767231595) += (0,0.007425543475062746) -= (0,0.007746525814329885)
(Social,0.5060244756989379) += (0,0.00862992954522357) -= (0,0.00859438033560228)
(All,0.5887940258624854) += (0,0.004103442535799973) -= (0,0.0041157209528478456)
          };
      \addplot[style={fill=lightgray},error bars/.cd, y dir=both, y explicit,error bar style=black]
           coordinates {
(Trauma,0.6378865787475075) += (0,0.0119576476105826) -= (0,0.012096314353275406)
(Anx,0.5922057266084959) += (0,0.004617719809004162) -= (0,0.004731922540469169)
(Compuls.,0.5856424758280658) += (0,0.017818249302948175) -= (0,0.017936496148185377)
(Cope,0.5887105654136473) += (0,0.0045978888128952455) -= (0,0.004486067568766683)
(Mood,0.6105143832329811) += (0,0.004169661426287274) -= (0,0.004238386512265935)
(Addict.,0.5408042001923349) += (0,0.0038984385211480177) -= (0,0.00395860140679527)
(Body,0.6155639543382515) += (0,0.01529421389797847) -= (0,0.01587817947763348)
(Neurodiv.,0.5992797132344616) += (0,0.0036899641469281264) -= (0,0.0036385952314014203)
(Health,0.6089407444992813) += (0,0.0027053359645056174) -= (0,0.002665433221403779)
(Social,0.6184805035707305) += (0,0.002843520445438563) -= (0,0.0028728095634809048)
(All,0.5969832369722273) += (0,0.0013587580377160124) -= (0,0.0013578362432092872)
          };
      \legend{MHPs, peers}
  \end{axis}
  \end{tikzpicture}
 \caption{LSM scores with 95\% confidence intervals calculated with non-parametric bootstrap resampling.}
 \label{fig:lsm}
\end{figure*}
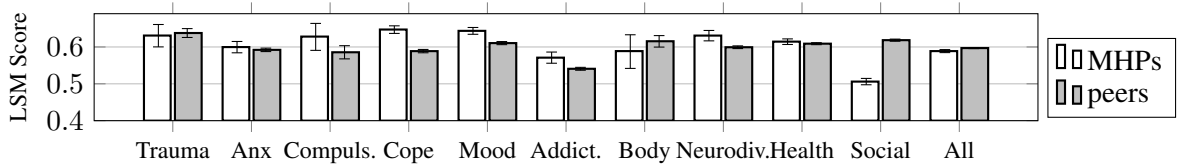
For a text sequence, we compute the percentage of words that belong to each of nine function-word categories $c$ from the LIWC lexicon, which include \textit{auxiliary verbs, articles, common adverbs, personal/impersonal pronouns, prepositions, negations, conjunctions,} and \textit{quantifiers}.
Then, we compute the LSM of each word category $c$ as shown in Equation \ref{eq:lsm} where $p$ represents \textit{post} and $r$ represents the response. The composite LSM score for $p$ and $r$ is the mean of all category LSM scores. For each thread, we separate the MHP and peer replies, and take the mean of all composite LSM scores. 

\begin{equation}\label{eq:lsm}
    LSM_c = 1 - \frac{abs(cat\%_p - cat\%_r)}{cat\%_p + cat\%_r + .0001}
\end{equation}

We compute these LSM scores over all data together as well as separately for each subreddit topic (named in Table~\ref{tab:subreddit_health_categories}). The resulting scores are shown in Figure~\ref{fig:lsm}. 

We observe LSM scores vary by topic, and most are similar for peers and MHPs or have overlapping confidence intervals. Compared to their LSMs in other topics, MHPs score lower in {\sc Social}, which covers broad social issues that are less specialized to health conditions than the others. However, peers have high LSMs in {\sc Social} relative to most other topics, and notably higher LSMs than the MHPs. Additionally, MHPs have higher LSMs than those of peers and relative to their own in communities that cover topics of specific compulsive, mood, and neurodevelopmental disorders ({\sc Compuls., Mood,} and {\sc Neurodiv.}), communities that orient toward counseling purposes ({\sc Cope}), or toward advice-seeking communities for health and social concerns ({\sc Health}). The influences in these results require further investigation, but a possible explanation could be that expert knowledge and experience may offer more benefit to specialized condition-related issues than to broader social issues.

\section{Language Modeling}

We further examine differences in word usage by building separate language models for MHPs and peers. We seek to identify language use that is indicative of one group or another by running the language model of one on the data of the other and analyzing words with high perplexity.
To run these experiments, we use the language model of \citet{merity2018analysis,merity2018regularizing}, which is a recent LSTM-based language model that achieved state-of-the-art performance by combining several regularization techniques.\footnote{\url{htts://github.com/salesforce/awd-lstm-lm}}

Our implementation uses a fixed vocabulary of 20,907 tokens for both the peer and MHP language models. This is determined by a minimum count of five across the set of posts from both groups. Each language model is trained for 50 epochs.\footnote{Validation set perplexities for expert and score groups: peer on peer: 44, peer on MHP: 52, MHP on peer: 91, MHP on MHP: 74, low on high: 39, low on low: 43, high on high: 50, high on low: 57. The difference in perplexity is due to the difference in volume of posts between groups. Runs in $\sim$2 min per epoch on a GeForce RTX 2080 Ti GPU.}

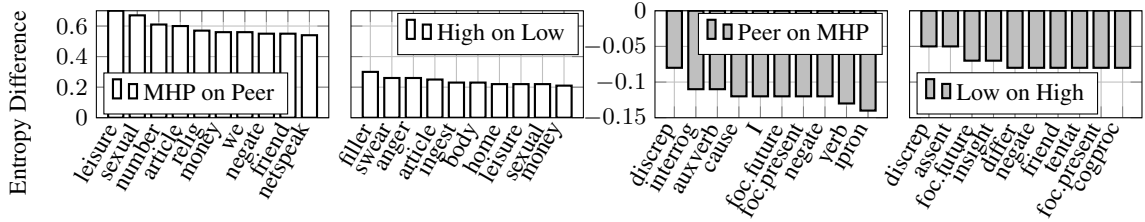
\begin{figure*}[t]
    \centering
    \small
    \begin{tikzpicture}
    \begin{axis}[
        yshift=0.0cm, xshift=0.0cm,
        width=0.29\linewidth,
        height=3cm,
        grid=both,
        symbolic x coords={leisure,sexual,number,article,relig,money,we,negate,friend,netspeak},
        xtick=data,
        legend style={font=\small},
        legend pos=south west,
        enlarge x limits=0.1,
        ymin=0.0, ymax=0.70,
        bar width=.2cm,
        x tick label style={rotate=60, anchor=east, font=\small},
        y tick label style={anchor=east, font=\small}]
        \addplot[ybar,fill=white,ybar legend] coordinates  {
            (leisure, 0.70)(sexual, 0.67)(number, 0.61)(article, 0.60)(relig, 0.57)(money, 0.56)(we, 0.56)(negate, 0.55)(friend, 0.55)(netspeak, 0.54)
        }; \addlegendentry{MHP on Peer}
    \end{axis}
    
    \begin{axis}[
        yshift=0.0cm, xshift=3.35cm,
        width=0.29\linewidth,
        height=3cm,
        grid=both,
        symbolic x coords={filler,swear,anger,article,ingest,body,home,leisure,sexual,money},
        xtick=data,
        enlarge x limits=0.1,
        ymin=0.0, ymax=0.70,
        legend style={font=\small},
        bar width=.2cm,
        x tick label style={rotate=60, anchor=east, font=\small},
        yticklabels={,,}
        ]
        \addplot[ybar,fill=white,ybar legend] coordinates  {
        	(filler, 0.30)(swear, 0.26)(anger, 0.26)(article, 0.25)(ingest, 0.23)(body, 0.23)(home, 0.22)(leisure, 0.22)(sexual, 0.22)(money, 0.21)
        }; \addlegendentry{High on Low}
    \end{axis}
    
    \begin{axis}[
        yshift=0cm, xshift=7.35cm,
        width=0.29\linewidth,
        height=3cm,
        grid=both,
        symbolic x coords={discrep,interrog,auxverb,cause,I,foc.future,foc.present,negate,verb,ipron},
        xtick=data,
        enlarge x limits=0.1,
        ymin=-0.15, ymax=0.0,
        legend style={font=\small},
        bar width=.2cm,
        x tick label style={rotate=60, anchor=east, font=\small},
        yticklabel style={
                /pgf/number format/fixed,
                /pgf/number format/precision=5,
                font=\small
        },
        scaled y ticks=false
        ]
        \addplot[ybar,fill=lightgray,ybar legend] coordinates  {
            (discrep, -0.08)(interrog, -0.11)(auxverb, -0.11)(cause, -0.12)(I, -0.12)(foc.future, -0.12)(foc.present, -0.12)(negate, -0.12)(verb, -0.13)(ipron, -0.14)
        }; \addlegendentry{Peer on MHP}
    \end{axis}
    
    \begin{axis}[
        yshift=0cm, xshift=10.7cm,
        width=0.29\linewidth,
        height=3cm,
        grid=both,
        symbolic x coords={discrep,assent,foc.future,insight,differ,negate,friend,tentat,foc.present,cogproc},
        xtick=data,
        legend style={font=\small},
        legend pos=south west,
        enlarge x limits=0.1,
        ymin=-0.15, ymax=0.0,
        bar width=.2cm,
        x tick label style={rotate=60, anchor=east, font=\small},
        yticklabels={,,}
        ]
        \addplot[ybar,fill=lightgray,ybar legend] coordinates  {
        	(discrep, -0.05)(assent, -0.05)(foc.future, -0.07)(insight, -0.07)(differ, -0.08)(negate, -0.08)(friend, -0.08)(tentat, -0.08)(foc.present, -0.08)(cogproc, -0.08)
        }; \addlegendentry{Low on High}
    \end{axis}
    
    \node[align=center,rotate=90,font=\small] at (-1.0cm, 0.2cm) {Entropy Difference};
    \end{tikzpicture}
    \caption{Entropy differences for LIWC word categories when running both language models on one group's data. High entropy scores on one dataset indicate word types that are harder for the opposite group's model to predict.}
    \label{fig:ent_diff_charts}
\end{figure*}

We use the language model trained on MHP data to find words with high entropy in peer data and vice versa. Since we are concerned with the \textit{difference} in predictability of words between the MHP and peer language models, we subtract the entropy given by the model trained on that data from the entropy assigned by the model that was not trained on that data. In other words, to find words difficult to predict in B's data, we subtract each word's entropy calculated by the model trained on B from the entropy calculated by the model trained on A as follows, for a set of words, X:

\begin{equation}
    E_{A,B} = -\frac{1}{\vert X \vert} \sum_{x \in X} log(p_{A}(x)) - log(p_{B}(x))
\end{equation}

If we calculate the entropy difference for each LIWC category and for each assignment of the MHP and peer groups to A and B, we find the highest differences for each category shown in the first and third plots of Figure~\ref{fig:ent_diff_charts}. We find highest entropy scores for words relating to \textit{leisure}, \textit{sex}, and \textit{numbers} when running the MHP language model on peer data.
Likewise, when running the peer model on MHP data, the category of \textit{discrepancy} contains words whose accuracy is improved the least by the peer model, again showing that these words are more indicative of the MHP group.

We perform a similar analysis, creating a language model for posts which have the highest score (or tied for highest) and another model for all other posts. We measure entropy differences and show the highest scoring categories for each group in the second and fourth plots. Some of the categories indicative of MHP language are also indicative of higher scoring posts; \textit{discrepancy}, \textit{present} and \textit{future} words, and \textit{negation} words, while other categories like \textit{assent} and \textit{insight} words are more dominant in higher scoring posts. The lower scoring posts have the highest entropy differences for some types of words in the peer data, however, we also see that \textit{filler}, \textit{anger}, and \textit{swear} words had the highest entropy differences for the low scoring group. Qualitative example sentences with word-level entropy and LIWC annotations are shown in the appendix in Table~\ref{tab:examples_mhp_data}.

\section{Discussion and Future Work}

In comparing linguistic aspects of MHP and peer comments, we find MHP tendencies align with established counseling principles and findings in counselor behaviors from recent literature. In particular, they align in the use of words that increase the likelihood of desired perspective shifts associated with clients feeling better after text counseling sessions (RQ2)~\cite{althoff2016large}. We also found unique differences in the behavior of MHPs as compared to peers in how they respond to information seekers (RQ1).
Although, comments by peers that prompt support-seeker replies also make use of similar word categories to MHPs, which shows that comparing MHPs to peers can offer insight into peer interactions as well.

It is important to note that our analyses rely heavily on the LIWC lexicon. While LIWC and other lexicons can help uncover variational language across groups at an exploratory stage, their use alone does not explain why variations are present. Certain limitations of LIWC are clear, such as when certain words that occur in multiple categories misleadingly boost the prominence of the categories equally. \citet{kross2019does}'s and \citet{jaidka2020estimating}'s studies have also demonstrated limitations of the use of LIWC when working with word counts to correlate with well-being metrics and an individual’s emotional state. We utilize LIWC to understand linguistic behavior differences in conversations with peers and MHPs rather than to evaluate the emotional or mental health state of individuals; however, it is important to consider how these limitations could pertain to our interpretations of their differences, especially as we explore them more deeply in future work. In our study, we explore the patterns we find in the context of previous findings from related literature such as \cite{althoff2016large} and \cite{nobles2020examining}, however it warrants another study into nuanced aspects of the patterns to infer their social functions in support seeking forums in particular.

Although our findings align MHP behaviors with certain counselor behaviors associated with positive outcomes, our analyses do not support claims that MHP behaviors are more beneficial to individuals seeking support; rather, we have shown that the general tendencies of MHPs are in accordance with principles and behaviors demonstrated by counselors in other settings. Understanding the outcomes of these interactions for individual support seekers remains as an area for future work, which could employ surveying methods from prior work to  measure perceived empathy in online communities~\cite{nambisan2011information}. Our dataset also enables investigations
into whether support-seekers exhibit perspective shifts in interacting with MHPs or peers, and what MHP and peer tendencies are associated with these perspective shifts.

Another direction for future work could focus on modeling social media-specific engagement patterns of MHP and peer interactions.
Prior work developed a model that accounts for variables indicating the level of attention threads receive (i.e., thread lengths and number of unique commenters), and variables indicating the degree of interaction between posters and commenters (i.e., time between responses and whether the poster replies to commenters), and used this model to study peer-to-peer interactions in online mental health platforms~\cite{sharma2020engagement}; this approach may enable studying supportive interactions in megathreads and threads involving back-and-forth dialog between two or more parties.

More questions arise if we consider MHP tenure and specific domain of expertise (e.g., specializations, licenses, academic degrees). Prior work that studied longitudinal changes in counselor linguistic behaviors indicated that systematic changes occur over time as counselors develop personal styles that are more distinct from other counselors and exhibit more diversity across interactions~\cite{zhang2019finding}. Future work could model the language longitudinally for MHPs and peers that have longer-term histories of participating in mental health forums to investigate whether systematic changes occur online as well, and if so, whether they reflect similar changes found in prior work.

\section{Limitations and Ethical Considerations}

A number of unknowns exist in what we are able to extract from Reddit. For instance, we do not know if users that do not use flairs are mental health professionals. We assume that those who have used the MHP flairs are MHPs and those that have not used them are peers. Additionally, we have grouped all MHP flairs into one group for our analysis, though a more nuanced analysis based on particular professional roles (e.g., psychologists, psychiatrists, social workers) and specializations (e.g., motivational interviewing, cognitive behavioral therapy, family \& marriage counseling) may reveal additional trends. Prior work found that disclosing credentials has impacts on engagements that vary by subreddit and linguistic patterns associated with different experience levels and expert domains~\cite{park2020trust}, thus the effects of disclosing MHP credentials when responding to support-seekers should be investigated.

A classifier or language model used to distinguish between MHPs and peers or to generate the language of either could have negative implications. A generative model that provides feedback to users could generate language that is harmful for those seeking help. Our work could be used to devise a tool to train counselors, however we do not have a direct measure of what type of responses are helpful or meaningful. In such an application, there is potential to reinforce harmful behaviors due to the inaccuracy of our models. Future studies are needed to determine how to best design a tool to train counselors and how models derived from corpora such as ours correspond to advice that patients find useful.

\section{Conclusion}

As the role of social networks is becoming more critical in how people seek health-information, it is important to understand their broader implications to health communication and how health experts can engage to promote the soundest information and offer support to their vulnerable users. By elucidating techniques employed by mental health professionals in their interactions with support-seekers in mental health forums, we have contributed insights toward the broader research direction of understanding how health experts currently engage with these platforms. With evidence that MHP linguistic behaviors associate with further engagement with support-seekers and that these same behaviors are associated with positive counseling conversation outcomes, we have shown that analyzing MHP behavior is a promising direction for better understanding online interaction outcomes, which can further inform forum design and moderation, and expert health promotion efforts. 

The code used for our experiments and analyses, and the post ids in our dataset can be accessed at \url{https://github.com/MichiganNLP/MHP-and-Peers-Reddit}.

\section*{Acknowledgments} 
This material is based in part upon work supported by the Precision Health initiative at the University of Michigan, by the National Science Foundation (grant \#1815291), and by the John Templeton Foundation (grant \#61156). Any opinions, findings, and conclusions or recommendations expressed in this material are those of the author and do not necessarily reflect the views of the Precision Health initiative, the National Science Foundation, or John Templeton Foundation.

\bibliographystyle{acl_natbib}
\bibliography{acl2021}

\begin{thebibliography}{48}
\expandafter\ifx\csname natexlab\endcsname\relax\def\natexlab#1{#1}\fi

\bibitem[{Althoff et~al.(2016)Althoff, Clark, and Leskovec}]{althoff2016large}
Tim Althoff, Kevin Clark, and Jure Leskovec. 2016.
\newblock Large-scale analysis of counseling conversations: An application of
  natural language processing to mental health.
\newblock \emph{Transactions of the Association for Computational Linguistics},
  4:463--476.

\bibitem[{Amir et~al.(2019)Amir, Dredze, and Ayers}]{amir2019mental}
Silvio Amir, Mark Dredze, and John~W Ayers. 2019.
\newblock Mental health surveillance over social media with digital cohorts.
\newblock In \emph{Proceedings of the Sixth Workshop on Computational
  Linguistics and Clinical Psychology}, pages 114--120.

\bibitem[{Baltadzhieva and Chrupa{\l}a(2015)}]{baltadzhieva-chrupala-2015}
Antoaneta Baltadzhieva and Grzegorz Chrupa{\l}a. 2015.
\newblock \href {https://www.aclweb.org/anthology/R15-1005} {Predicting the
  quality of questions on {S}tackoverflow}.
\newblock In \emph{Proceedings of the International Conference Recent Advances
  in Natural Language Processing}, pages 32--40, Hissar, Bulgaria. INCOMA Ltd.
  Shoumen, Bulgaria.

\bibitem[{Campbell et~al.(2016)Campbell, Aragon, Davis, Evans, Evans, and
  Randall}]{Campbell16}
Julie Campbell, Cecilia Aragon, Katie Davis, Sarah Evans, Abigail Evans, and
  David Randall. 2016.
\newblock \href {https://doi.org/10.1145/2818048.2819934} {Thousands of
  positive reviews: Distributed mentoring in online fan communities}.
\newblock In \emph{Proceedings of the 19th ACM Conference on Computer-Supported
  Cooperative Work \& Social Computing}, CSCW '16, page 691–704, New York,
  NY, USA. Association for Computing Machinery.

\bibitem[{Campbell and Pennebaker(2003)}]{campbell2003secret}
R~Sherlock Campbell and James~W Pennebaker. 2003.
\newblock The secret life of pronouns: Flexibility in writing style and
  physical health.
\newblock \emph{Psychological science}, 14(1):60--65.

\bibitem[{Chou et~al.(2009)Chou, Hunt, Beckjord, Moser, and
  Hesse}]{chou2009social}
Wen-Ying~Sylvia Chou, Yvonne~M Hunt, Ellen~B Beckjord, Richard~P Moser, and
  Bradford~W Hesse. 2009.
\newblock Social media use in the united states: implications for health
  communication.
\newblock \emph{Journal of medical Internet research}, 11(4):e48.

\bibitem[{De~Choudhury and De(2014)}]{de2014mental}
Munmun De~Choudhury and Sushovan De. 2014.
\newblock Mental health discourse on reddit: Self-disclosure, social support,
  and anonymity.
\newblock In \emph{Proceedings of the International AAAI Conference on Web and
  Social Media}.

\bibitem[{Dom{\'\i}nguez and Sapi{\~n}a(2015)}]{dominguez2015pediatric}
Mart{\'\i} Dom{\'\i}nguez and Luc{\'\i}a Sapi{\~n}a. 2015.
\newblock Pediatric cancer and the internet: exploring the gap in
  doctor-parents communication.
\newblock \emph{Journal of Cancer Education}, 30(1):145--151.

\bibitem[{Dror et~al.(2018)Dror, Baumer, Shlomov, and
  Reichart}]{dror-etal-2018-hitchhikers2}
Rotem Dror, Gili Baumer, Segev Shlomov, and Roi Reichart. 2018.
\newblock \href {https://doi.org/10.18653/v1/P18-1128} {The hitchhiker{'}s
  guide to testing statistical significance in natural language processing}.
\newblock In \emph{Proceedings of the 56th Annual Meeting of the Association
  for Computational Linguistics (Volume 1: Long Papers)}, pages 1383--1392,
  Melbourne, Australia. Association for Computational Linguistics.

\bibitem[{Gage-Bouchard et~al.(2018)Gage-Bouchard, LaValley, Warunek, Beaupin,
  and Mollica}]{gage2018cancer}
Elizabeth~A Gage-Bouchard, Susan LaValley, Molli Warunek, Lynda~Kwon Beaupin,
  and Michelle Mollica. 2018.
\newblock Is cancer information exchanged on social media scientifically
  accurate?
\newblock \emph{Journal of cancer Education}, 33(6):1328--1332.

\bibitem[{Gonzales et~al.(2010)Gonzales, Hancock, and
  Pennebaker}]{gonzales2010language}
Amy~L Gonzales, Jeffrey~T Hancock, and James~W Pennebaker. 2010.
\newblock Language style matching as a predictor of social dynamics in small
  groups.
\newblock \emph{Communication Research}, 37(1):3--19.

\bibitem[{Griffiths et~al.(2012)Griffiths, Cave, Boardman, Ren, Pawlikowska,
  Ball, Clarke, and Cohen}]{griffiths2012social}
Frances Griffiths, Jonathan Cave, Felicity Boardman, Justin Ren, Teresa
  Pawlikowska, Robin Ball, Aileen Clarke, and Alan Cohen. 2012.
\newblock Social networks--the future for health care delivery.
\newblock \emph{Social science \& medicine}, 75(12):2233--2241.

\bibitem[{Hodges et~al.(2010)Hodges, Kiel, Kramer, Veach, and
  Villanueva}]{hodges2010giving}
Sara~D Hodges, Kristi~J Kiel, Adam~DI Kramer, Darya Veach, and B~Renee
  Villanueva. 2010.
\newblock Giving birth to empathy: The effects of similar experience on
  empathic accuracy, empathic concern, and perceived empathy.
\newblock \emph{Personality and Social Psychology Bulletin}, 36(3):398--409.

\bibitem[{Ireland and Pennebaker(2010)}]{ireland2010language}
Molly~E Ireland and James~W Pennebaker. 2010.
\newblock Language style matching in writing: Synchrony in essays,
  correspondence, and poetry.
\newblock \emph{Journal of personality and social psychology}, 99(3):549.

\bibitem[{Jaidka et~al.(2020)Jaidka, Giorgi, Schwartz, Kern, Ungar, and
  Eichstaedt}]{jaidka2020estimating}
Kokil Jaidka, Salvatore Giorgi, H~Andrew Schwartz, Margaret~L Kern, Lyle~H
  Ungar, and Johannes~C Eichstaedt. 2020.
\newblock Estimating geographic subjective well-being from twitter: A
  comparison of dictionary and data-driven language methods.
\newblock \emph{Proceedings of the National Academy of Sciences},
  117(19):10165--10171.

\bibitem[{Kross et~al.(2019)Kross, Verduyn, Boyer, Drake, Gainsburg, Vickers,
  Ybarra, and Jonides}]{kross2019does}
Ethan Kross, Philippe Verduyn, Margaret Boyer, Brittany Drake, Izzy Gainsburg,
  Brian Vickers, Oscar Ybarra, and John Jonides. 2019.
\newblock Does counting emotion words on online social networks provide a
  window into people’s subjective experience of emotion? a case study on
  facebook.
\newblock \emph{Emotion}, 19(1):97.

\bibitem[{Merity et~al.(2018{\natexlab{a}})Merity, Keskar, and
  Socher}]{merity2018analysis}
Stephen Merity, Nitish~Shirish Keskar, and Richard Socher. 2018{\natexlab{a}}.
\newblock An analysis of neural language modeling at multiple scales.
\newblock \emph{arXiv preprint arXiv:1803.08240}.

\bibitem[{Merity et~al.(2018{\natexlab{b}})Merity, Keskar, and
  Socher}]{merity2018regularizing}
Stephen Merity, Nitish~Shirish Keskar, and Richard Socher. 2018{\natexlab{b}}.
\newblock \href {https://openreview.net/forum?id=SyyGPP0TZ} {Regularizing and
  optimizing {LSTM} language models}.
\newblock In \emph{International Conference on Learning Representations}.

\bibitem[{Mihalcea and Pulman(2009)}]{mihalcea2009linguistic}
Rada Mihalcea and Stephen Pulman. 2009.
\newblock Linguistic ethnography: Identifying dominant word classes in text.
\newblock In \emph{International Conference on Intelligent Text Processing and
  Computational Linguistics}, pages 594--602. Springer.

\bibitem[{Miller and Rollnick(2012)}]{miller2012motivational}
William~R Miller and Stephen Rollnick. 2012.
\newblock \emph{Motivational interviewing: Helping people change}.
\newblock Guilford press.

\bibitem[{Nambisan(2011)}]{nambisan2011information}
Priya Nambisan. 2011.
\newblock Information seeking and social support in online health communities:
  impact on patients' perceived empathy.
\newblock \emph{Journal of the American Medical Informatics Association},
  18(3):298--304.

\bibitem[{Nobles et~al.(2018)Nobles, Dreisbach, Keim-Malpass, and
  Barnes}]{nobles2018std}
Alicia Nobles, Caitlin Dreisbach, Jessica Keim-Malpass, and Laura Barnes. 2018.
\newblock " is this an std? please help!": Online information seeking for
  sexually transmitted diseases on reddit.
\newblock In \emph{Proceedings of the International AAAI Conference on Web and
  Social Media}.

\bibitem[{Nobles et~al.(2019)Nobles, Leas, Althouse, Dredze, Longhurst, Smith,
  and Ayers}]{nobles2019requests}
Alicia~L Nobles, Eric~C Leas, Benjamin~M Althouse, Mark Dredze, Christopher~A
  Longhurst, Davey~M Smith, and John~W Ayers. 2019.
\newblock Requests for diagnoses of sexually transmitted diseases on a social
  media platform.
\newblock \emph{Jama}, 322(17):1712--1713.

\bibitem[{Nobles et~al.(2020)Nobles, Leas, Dredze, and
  Ayers}]{nobles2020examining}
Alicia~L Nobles, Eric~C Leas, Mark Dredze, and John~W Ayers. 2020.
\newblock Examining peer-to-peer and patient-provider interactions on a social
  media community facilitating ask the doctor services.
\newblock In \emph{Proceedings of the International AAAI Conference on Web and
  Social Media}, volume~14, pages 464--475.

\bibitem[{Park and Conway(2017)}]{park2017tracking}
Albert Park and Mike Conway. 2017.
\newblock Tracking health related discussions on reddit for public health
  applications.
\newblock In \emph{AMIA Annual Symposium Proceedings}, volume 2017, page 1362.
  American Medical Informatics Association.

\bibitem[{Park et~al.(2020)Park, Kwak, Song, and Cha}]{park2020trust}
Kunwoo Park, Haewoon Kwak, Hyunho Song, and Meeyoung Cha. 2020.
\newblock “trust me, i have a ph. d.”: A propensity score analysis on the
  halo effect of disclosing one's offline social status in online communities.
\newblock In \emph{Proceedings of the International AAAI Conference on Web and
  Social Media}, volume~14, pages 534--544.

\bibitem[{Pennebaker et~al.(2015)Pennebaker, Booth, Boyd, and
  Francis}]{pennebakerlinguistic}
James~W Pennebaker, Roger~J Booth, Ryan~L Boyd, and Martha~E Francis. 2015.
\newblock \href {www.LIWC.net} {Linguistic inquiry and word count: Liwc2015}.

\bibitem[{P{\'e}rez-Rosas et~al.(2018)P{\'e}rez-Rosas, Sun, Li, Wang, Resnicow,
  and Mihalcea}]{perez2018analyzing}
Ver{\'o}nica P{\'e}rez-Rosas, Xuetong Sun, Christy Li, Yuchen Wang, Kenneth
  Resnicow, and Rada Mihalcea. 2018.
\newblock Analyzing the quality of counseling conversations: the tell-tale
  signs of high-quality counseling.
\newblock In \emph{Proceedings of the Eleventh International Conference on
  Language Resources and Evaluation (LREC 2018)}.

\bibitem[{P{\'e}rez-Rosas et~al.(2019)P{\'e}rez-Rosas, Wu, Resnicow, and
  Mihalcea}]{perez2019makes}
Ver{\'o}nica P{\'e}rez-Rosas, Xinyi Wu, Kenneth Resnicow, and Rada Mihalcea.
  2019.
\newblock What makes a good counselor? learning to distinguish between
  high-quality and low-quality counseling conversations.
\newblock In \emph{Proceedings of the 57th Annual Meeting of the Association
  for Computational Linguistics}, pages 926--935.

\bibitem[{Poland et~al.(2011)Poland, Jacobson et~al.}]{poland2011age}
Gregory~A Poland, Robert~M Jacobson, et~al. 2011.
\newblock The age-old struggle against the antivaccinationists.
\newblock \emph{N Engl J Med}, 364(2):97--9.

\bibitem[{{Procaci} et~al.(2017){Procaci}, {Siqueira}, {Nunes}, and
  {Nurmikko-Fuller}}]{Procaci17}
T.~B. {Procaci}, S.~W.~M. {Siqueira}, B.~P. {Nunes}, and T.~{Nurmikko-Fuller}.
  2017.
\newblock \href {https://doi.org/10.1109/ICALT.2017.56} {Modelling experts
  behaviour in q a communities to predict worthy discussions}.
\newblock In \emph{2017 IEEE 17th International Conference on Advanced Learning
  Technologies (ICALT)}, pages 291--295.

\bibitem[{Pyszczynski and Greenberg(1987)}]{pyszczynski1987self}
Tom Pyszczynski and Jeff Greenberg. 1987.
\newblock Self-regulatory perseveration and the depressive self-focusing style:
  a self-awareness theory of reactive depression.
\newblock \emph{Psychological bulletin}, 102(1):122.

\bibitem[{Pyszczynski et~al.(1987)Pyszczynski, Holt, and
  Greenberg}]{pyszczynski1987depression}
Tom Pyszczynski, Kathleen Holt, and Jeff Greenberg. 1987.
\newblock Depression, self-focused attention, and expectancies for positive and
  negative future life events for self and others.
\newblock \emph{Journal of personality and social psychology}, 52(5):994.

\bibitem[{Record et~al.(2018)Record, Silberman, Santiago, and
  Ham}]{record2018sought}
Rachael~A Record, Will~R Silberman, Joshua~E Santiago, and Taewook Ham. 2018.
\newblock I sought it, i reddit: Examining health information engagement
  behaviors among reddit users.
\newblock \emph{Journal of Health Communication}, 23(5):470--476.

\bibitem[{Rheingold(1993)}]{rheingold1993virtual}
Howard Rheingold. 1993.
\newblock \emph{The virtual community: Homesteading on the electronic
  frontier}, volume~32.
\newblock Addison-Wesley Reading, MA.

\bibitem[{Rude et~al.(2004)Rude, Gortner, and Pennebaker}]{rude2004language}
Stephanie~S Rude, Eva-Maria Gortner, and James~W Pennebaker. 2004.
\newblock Language use of depressed and depression-vulnerable college students.
\newblock \emph{Cognition and Emotion}, 18(8):1121--1133.

\bibitem[{Sharma et~al.(2020)Sharma, Choudhury, Althoff, and
  Sharma}]{sharma2020engagement}
Ashish Sharma, Monojit Choudhury, Tim Althoff, and Amit Sharma. 2020.
\newblock Engagement patterns of peer-to-peer interactions on mental health
  platforms.
\newblock In \emph{Proceedings of the International AAAI Conference on Web and
  Social Media}, volume~14, pages 614--625.

\bibitem[{Sharma and De~Choudhury(2018)}]{sharma2018support}
Eva Sharma and Munmun De~Choudhury. 2018.
\newblock \href {https://doi.org/10.1145/3173574.3174215} {Mental health
  support and its relationship to linguistic accommodation in online
  communities}.
\newblock In \emph{Proceedings of the 2018 CHI Conference on Human Factors in
  Computing Systems}, CHI '18, page 1–13, New York, NY, USA. Association for
  Computing Machinery.

\bibitem[{Strapparava and Valitutti(2004)}]{wordnet-affect}
Carlo Strapparava and Alessandro Valitutti. 2004.
\newblock Wordnet-affect: an affective extension of wordnet.
\newblock In \emph{Proceedings of the 4th International Conference on Language
  Resources and Evaluation (LREC 2004)}, pages 1083--1086.

\bibitem[{Tausczik and Pennebaker(2010)}]{tausczik2010psychological}
Yla~R Tausczik and James~W Pennebaker. 2010.
\newblock The psychological meaning of words: Liwc and computerized text
  analysis methods.
\newblock \emph{Journal of language and social psychology}, 29(1):24--54.

\bibitem[{Tomprou et~al.(2019)Tomprou, Dabbish, Kraut, and Liu}]{Tomprou19}
Maria Tomprou, Laura Dabbish, Robert~E. Kraut, and Fannie Liu. 2019.
\newblock \href {https://doi.org/10.1145/3290605.3300883} {Career mentoring in
  online communities: Seeking and receiving advice from an online community}.
\newblock In \emph{Proceedings of the 2019 CHI Conference on Human Factors in
  Computing Systems}, CHI '19, page 1–12, New York, NY, USA. Association for
  Computing Machinery.

\bibitem[{Wang et~al.(2014)Wang, Li, Ferguson, and
  Zhai}]{wang2014sideeffectptm}
Sheng Wang, Yanen Li, Duncan Ferguson, and Chengxiang Zhai. 2014.
\newblock Sideeffectptm: An unsupervised topic model to mine adverse drug
  reactions from health forums.
\newblock In \emph{Proceedings of the 5th ACM conference on bioinformatics,
  computational biology, and health informatics}, pages 321--330.

\bibitem[{Wang et~al.(2015)Wang, Kraut, and Levine}]{wang2015eliciting}
Yi-Chia Wang, Robert~E Kraut, and John~M Levine. 2015.
\newblock Eliciting and receiving online support: using computer-aided content
  analysis to examine the dynamics of online social support.
\newblock \emph{Journal of medical Internet research}, 17(4):e99.

\bibitem[{Wang and Jurgens(2018)}]{wang2018s}
Zijian Wang and David Jurgens. 2018.
\newblock It’s going to be okay: Measuring access to support in online
  communities.
\newblock In \emph{Proceedings of the 2018 Conference on Empirical Methods in
  Natural Language Processing}, pages 33--45.

\bibitem[{Weintraub(1981)}]{weintraub1981verbal}
Walter Weintraub. 1981.
\newblock \emph{Verbal behavior: Adaptation and psychopathology}.
\newblock Springer Publishing Company New York.

\bibitem[{Welbourne et~al.(2009)Welbourne, Blanchard, and
  Boughton}]{welbourne2009supportive}
Jennifer~L Welbourne, Anita~L Blanchard, and Marla~D Boughton. 2009.
\newblock Supportive communication, sense of virtual community and health
  outcomes in online infertility groups.
\newblock In \emph{Proceedings of the fourth international conference on
  Communities and technologies}, pages 31--40.

\bibitem[{Zhang et~al.(2019)Zhang, Filbin, Morrison, Weiser, and
  Danescu-Niculescu-Mizil}]{zhang2019finding}
Justine Zhang, Robert Filbin, Christine Morrison, Jaclyn Weiser, and Cristian
  Danescu-Niculescu-Mizil. 2019.
\newblock Finding your voice: The linguistic development of mental health
  counselors.
\newblock In \emph{Proceedings of the 57th Annual Meeting of the Association
  for Computational Linguistics}, pages 936--947.

\bibitem[{Zhou and Jurgens(2020)}]{zhou2020condolences}
Naitian Zhou and David Jurgens. 2020.
\newblock Condolences and empathy in online communities.
\newblock In \emph{Proceedings of the 2020 Conference on Empirical Methods in
  Natural Language Processing (EMNLP)}, pages 609--626.

\end{thebibliography}

\appendix

\section*{Appendix}

\section{Flairs}\label{sec:appendix_flairs}

Rules regarding flair credibility from r/psychotherapy:

``\textbf{2. Only posts and comments from those providing therapy in a licensed capacity allowed.} No comments/posts from anyone who is not providing therapy in a licensed capacity. This includes students who are not yet practicing therapy (e.g., undergraduate or graduate students who haven't had their first practica experience) or if you have left the field for another field, this is not the place for you to post/comment. There is an exception to this rule for posting in our Career and Education Megathread. Accurate user flair is required for all posts, and strongly encouraged for comments." 

``\textbf{9. Falsely posing as a therapist} If you post in this subreddit, the assumption is made that you are a therapist. Users that falsely post as if they were a therapist will be permanently banned. Claiming that you didn’t say you were a therapist is not an argument against this rule. Users may be asked to submit proof of their status as a practicing therapist to appeal a ban."

\section{Data}\label{sec:appendix_datacollection}

We used the PushShift API for the first pass of obtaining mental health posts and comments, and the MHP flairs. After extracting the IDs of posts where MHPs commented, we obtained the fully structured comment sections using open sourced code from \url{https://github.com/saucecode/reddit-thread-ripper}. The numbers of posts in our dataset for each subreddit are shown in Table~\ref{tab:appendix_subbreddits_and_comments}.

\begin{table*}[t]
    \centering
    \small
    \begin{tabular}{lrlrlr}
\toprule
AskDocs & 21025 & relationship\_advice & 16061 & stopdrinking & 10170 \\
ADHD & 9093 & offmychest & 5076 & mentalhealth & 4486 \\
socialskills & 4113 & BPD & 3570 & depression & 3235 \\
Anxiety & 2956 & aspergers & 2703 & Advice & 2493 \\
askatherapist & 2120 & PCOS & 1819 & alcoholicsanonymous & 1498 \\
leaves & 1452 & SuicideWatch & 1092 & REDDITORSINRECOVERY & 977 \\
needadvice & 892 & ptsd & 704 & NoFap & 465 \\
OCD & 411 & socialanxiety & 400 & BipolarReddit & 355 \\
GetMotivated & 354 & alcoholism & 350 & cripplingalcoholism & 335 \\
emetophobia & 297 & bulimia & 249 & mentalillness & 246 \\
nosurf & 224 & EOOD & 208 & depression\_help & 193 \\
EatingDisorders & 170 & schizophrenia & 167 & MMFB & 159 \\
AlAnon & 139 & disability & 127 & fuckeatingdisorders & 119 \\
Antipsychiatry & 116 & MadOver30 & 114 & quittingkratom & 114 \\
addiction & 111 & GFD & 109 & CompulsiveSkinPicking & 108 \\
Needafriend & 106 & dbtselfhelp & 99 & rapecounseling & 93 \\
stopsmoking & 89 & selfhelp & 87 & ForeverAlone & 81 \\
getting\_over\_it & 72 & BodyAcceptance & 54 & Anger & 50 \\
traumatoolbox & 50 & selfharm & 47 & TwoXADHD & 40 \\
survivorsofabuse & 40 & dpdr & 38 & rape & 36 \\
Tourettes & 34 & HealthAnxiety & 26 & schizoaffective & 25 \\
Anxietyhelp & 25 & eating\_disorders & 20 & domesticviolence & 17 \\
neurodiversity & 13 & helpmecope & 12 & StopSelfHarm & 12 \\
sad & 11 & AtheistTwelveSteppers & 10 & Trichsters & 6 \\
MenGetRapedToo & 5 & ARFID & 5 & whatsbotheringyou & 3 \\
DysmorphicDisorder & 1 & OCPD & 1 &  &  \\
\bottomrule

    \end{tabular}
    \caption{The number of comments in each subreddit of our dataset.}
    \label{tab:appendix_subbreddits_and_comments}
\end{table*}

\begin{table*}
    \centering
    \small
    \begin{tabular}{p{0.3\linewidth}p{0.6\linewidth}}
\toprule
Category & Subreddits \\
\midrule
Trauma \& Abuse (Trauma) & r/Anger, r/survivorsofabuse, r/domesticviolence, r/ptsd, r/rapecounseling, r/selfharm, r/StopSelfHarm, r/traumatoolbox, r/rape, r/MenGetRapedToo \\ 
Psychosis \& Anxiety (Anx) & r/Anxiety, r/socialanxiety, r/Anxietyhelp, r/HealthAnxiety, r/BPD, r/dpdr, r/schizophrenia, r/schizoaffective, r/emetophobia \\ 
Compulsive Disorders (Compuls.) & r/CompulsiveSkinPicking, r/OCD, r/Trichsters, r/DysmorphicDisorder, r/OCPD \\ 
Coping \& Therapy (Cope) & r/getting\_over\_it, r/helpmecope, r/offmychest, r/MMFB, r/askatherapist, r/EOOD, r/dbtselfhelp, r/AlAnon, r/REDDITORSINRECOVERY, r/GetMotivated, r/Antipsychiatry, r/selfhelp \\ 
Mood Disorders (Mood) & r/depression, r/depression\_help, r/ForeverAlone, r/GFD, r/mentalhealth, r/SuicideWatch, r/sad, r/BipolarReddit \\ 
Addiction \& Impulse Control (Addict.) & r/stopdrinking, r/addiction, r/stopsmoking, r/leaves, r/alcoholism, r/cripplingalcoholism, r/quittingkratom, r/alcoholicsanonymous, r/NoFap \\ 
Eating \& Body (Body) & r/eating\_disorders, r/EatingDisorders, r/ARFID, r/fuckeatingdisorders, r/BodyAcceptance,  r/bulimia \\ 
Neurodevelopmental Disorders (Neurodiv.) & r/ADHD, r/aspergers, r/TwoXADHD \\ 
General (Health) & r/AskDocs, r/needadvice, r/Advice, r/mentalillness, r/neurodiversity, r/whatsbotheringyou, r/MadOver30 \\ 
Broad Social (Social) & r/socialskills, r/relationship\_advice, r/nosurf, r/Needafriend, r/AtheistTwelveSteppers, r/PCOS, r/disability, r/Tourettes \\ 
Overall & All \\ 
\bottomrule
    \end{tabular}
    \caption{Subreddit categories.}
    \label{tab:appendix_subreddit_categories}
\end{table*}

\section{Other}

Sample sentences illustrating relative entropies of words predicted by the peer language model on MHP data (top) and the MHP language model on peer data (bottom) are shown in Table~\ref{tab:examples_mhp_data}.

\begin{table*}
    \centering
    \small
    \begin{tabular}{lp{13.0cm}}
        \toprule
        \multicolumn{2}{c}{MHP Data} \\
        \midrule
        DISCREP & \noindent\parbox[c]{\hsize}{{\color{white}\hlc[blue!78]{the}} \autour{blue!100}{white}{problem} {\color{white}\hlc[blue!74]{with}} {\color{white}\hlc[blue!61]{psychiatric}} {\color{white}\hlc[blue!84]{research}} {\color{white}\hlc[blue!61]{is}} {\color{white}\hlc[blue!91]{the}} {\color{white}\hlc[blue!90]{relative}} {\color{black}\hlc[blue!0]{subjectivity}} {\color{white}\hlc[blue!77]{of}} {\color{white}\hlc[blue!74]{it}} {\color{white}\hlc[blue!76]{,}} {\color{white}\hlc[blue!99]{much}} {\color{white}\hlc[blue!88]{less}} {\color{black}\hlc[blue!49]{glamorous}} {\color{white}\hlc[blue!80]{outcomes}} {\color{white}\hlc[blue!86]{,}} {\color{white}\hlc[blue!74]{and}} {\color{white}\hlc[blue!88]{the}} \autour{blue!78}{white}{lack} {\color{white}\hlc[blue!92]{of}} {\color{black}\hlc[blue!48]{public}} {\color{white}\hlc[blue!73]{interest}} {\color{white}\hlc[blue!84]{despite}} {\color{white}\hlc[blue!73]{its}} {\color{white}\hlc[blue!74]{burden}} {\color{black}\hlc[blue!46]{on}} {\color{white}\hlc[blue!79]{society}} {\color{white}\hlc[blue!78]{.}}} \\
        INTERROG & \noindent\parbox[c]{\hsize}{\autour{blue!100}{white}{who} {\color{white}\hlc[blue!77]{diagnosed}} {\color{white}\hlc[blue!71]{you}} {\color{black}\hlc[blue!28]{with}} {\color{black}\hlc[blue!0]{spinal}} {\color{white}\hlc[blue!64]{issues}} {\color{white}\hlc[blue!77]{-}} \autour{blue!65}{white}{which} {\color{white}\hlc[blue!58]{might}} {\color{white}\hlc[blue!52]{show}} \autour{blue!88}{white}{how} {\color{white}\hlc[blue!73]{anxiety}} {\color{white}\hlc[blue!63]{affects}} {\color{black}\hlc[blue!35]{your}} {\color{black}\hlc[blue!39]{physical}} {\color{white}\hlc[blue!68]{health}} {\color{white}\hlc[blue!65]{.}}} \\
        AUXVERB & \noindent\parbox[c]{\hsize}{{\color{black}\hlc[blue!30]{that}} \autour{blue!23}{black}{being} {\color{black}\hlc[blue!23]{said}} {\color{black}\hlc[blue!42]{,}} {\color{black}\hlc[blue!6]{many}} {\color{black}\hlc[blue!0]{other}} {\color{black}\hlc[blue!27]{mental}} {\color{black}\hlc[blue!48]{health}} {\color{black}\hlc[blue!21]{concerns}} \autour{blue!100}{white}{have} {\color{white}\hlc[blue!50]{overlapping}} {\color{black}\hlc[blue!21]{symptoms}} {\color{black}\hlc[blue!36]{with}} {\color{black}\hlc[blue!12]{adhd}} {\color{black}\hlc[blue!36]{including}} {\color{black}\hlc[blue!26]{anxiety}} {\color{black}\hlc[blue!4]{and}} {\color{black}\hlc[blue!25]{depression}} {\color{black}\hlc[blue!23]{.}}} \\
        \midrule
        \multicolumn{2}{c}{Peer Data} \\
        \midrule
        LEISURE & \noindent\parbox[c]{\hsize}{{\color{black}\hlc[blue!22]{it}} {\color{black}\hlc[blue!17]{'s}} {\color{black}\hlc[blue!32]{like}} \autour{blue!93}{white}{jogging} {\color{black}\hlc[blue!19]{with}} {\color{black}\hlc[blue!0]{a}} {\color{black}\hlc[blue!33]{back}} {\color{white}\hlc[blue!62]{back}} {\color{black}\hlc[blue!3]{full}} {\color{black}\hlc[blue!21]{of}} {\color{white}\hlc[blue!100]{bricks}} {\color{black}\hlc[blue!11]{.}}} \\
        SEXUAL & \noindent\parbox[c]{\hsize}{{\color{black}\hlc[blue!22]{it}} {\color{black}\hlc[blue!20]{seems}} {\color{black}\hlc[blue!29]{like}} {\color{white}\hlc[blue!58]{everybody}} {\color{black}\hlc[blue!22]{here}} {\color{black}\hlc[blue!46]{is}} {\color{white}\hlc[blue!61]{desired}} \autour{blue!99}{white}{sexually} {\color{black}\hlc[blue!23]{so}} {\color{black}\hlc[blue!7]{that}} {\color{black}\hlc[blue!8]{must}} {\color{black}\hlc[blue!0]{mean}} {\color{black}\hlc[blue!26]{you}} {\color{black}\hlc[blue!18]{'re}} {\color{black}\hlc[blue!25]{doing}} {\color{black}\hlc[blue!32]{something}} {\color{white}\hlc[blue!57]{right}} {\color{black}\hlc[blue!16]{.}}} \\
        NUMBER & \noindent\parbox[c]{\hsize}{{\color{black}\hlc[blue!6]{when}} {\color{black}\hlc[blue!12]{it}} {\color{black}\hlc[blue!22]{came}} {\color{black}\hlc[blue!8]{time}} {\color{black}\hlc[blue!7]{for}} {\color{black}\hlc[blue!25]{homework}} {\color{black}\hlc[blue!19]{she}} {\color{black}\hlc[blue!6]{set}} {\color{black}\hlc[blue!10]{them}} {\color{black}\hlc[blue!10]{up}} {\color{black}\hlc[blue!15]{with}} \autour{blue!0}{black}{three} {\color{black}\hlc[blue!11]{things}} {\color{black}\hlc[blue!5]{to}} {\color{black}\hlc[blue!11]{do}} {\color{black}\hlc[blue!4]{(}} \autour{blue!4}{black}{one} {\color{black}\hlc[blue!9]{being}} {\color{black}\hlc[blue!0]{homework}} {\color{black}\hlc[blue!9]{)}} {\color{black}\hlc[blue!9]{and}} {\color{black}\hlc[blue!3]{had}} {\color{black}\hlc[blue!19]{them}} {\color{black}\hlc[blue!2]{switch}} {\color{black}\hlc[blue!35]{every}} \autour{blue!99}{white}{fifteen} {\color{black}\hlc[blue!10]{minutes}} {\color{black}\hlc[blue!6]{.}}} \\
        \bottomrule
    \end{tabular}
    \caption{Sample sentences from MHP data with relative entropy marked by highlight color (i.e. darker blue means higher entropy relative to other words in the sentence). All words in the given LIWC category are marked with a rounded rectangle.}
    \label{tab:examples_mhp_data}
\end{table*}

\end{document}